\documentclass[final]{cvpr}
\pdfoutput=1

\usepackage{times}
\usepackage{epsfig}
\usepackage{graphicx}
\usepackage{amsmath}
\usepackage{amssymb}
\usepackage{bbm}
\usepackage{makecell}
\usepackage{multirow}
\usepackage{multicol}
\usepackage[pagebackref=true,breaklinks=true,colorlinks,bookmarks=false]{hyperref}

\begin{document}

\begin{twocolumn}
%%%%%%%%% TITLE
\title{Indicative Image Retrieval: Turning Blackbox Learning into Grey}

% \author{Xu-Lu Zhang\\
% Sichuan University\\
% \and

\author{Xu-Lu Zhang\textsuperscript{1}, Zhen-Qun Yang\textsuperscript{2}, Hao Tian\textsuperscript{1}, Qing Li\textsuperscript{3}, Xiao-Yong Wei\textsuperscript{1,3,*}\\
\textsuperscript{1} Sichuan University, \textsuperscript{2} Chinese University of Hong Kong, \textsuperscript{3} Hong Kong Polytechnic Univeristy\\
{\textsuperscript{*}\tt\small  Corresponding cswei@scu.edu.cn, x1wei@polyu.edu.hk}
% \and
}

\maketitle

%%%%%%%%% ABSTRACT
\begin{abstract}
Deep learning became the game changer for image retrieval soon after it was introduced. It promotes the feature extraction (by representation learning) as the core of image retrieval, with the relevance/matching evaluation being degenerated into simple similarity metrics. 
In many applications, we need the matching evidence to be indicated rather than just have the ranked list (e.g., the locations of the target proteins/cells/lesions in medical images). It is like the matched words need to be highlighted in search engines.
However, this is not easy to implement without explicit relevance/matching modeling. The deep representation learning models are not feasible because of their blackbox nature.
In this paper, we revisit the importance of relevance/matching modeling in deep learning era with an indicative retrieval setting. The study shows that it is possible to skip the representation learning and model the matching evidence directly.
By removing the dependency on the pre-trained models, it has avoided a lot of related issues (e.g., the {domain gap} between classification and retrieval, the {detail-diffusion} caused by convolution, and so on).
More importantly, the study demonstrates that the matching can be explicitly modeled and backtracked later for generating the matching evidence indications. It can improve the explainability of deep inference.
Our method obtains a best performance in literature on both Oxford-5k and Paris-6k, and sets a new record of $97.77\%$ on Oxford-5k ($97.81\%$ on Paris-6k) without extracting any deep features.
\end{abstract}

%%%%%%%%% BODY TEXT
\section{Introduction}
Content-based Image Retrieval (CBIR) is generally modeled as a ranking problem based on the relevance/matching between the query and target images \cite{chen2021deep,zheng2017sift,smeulders2000content}. More specifically, the feature vectors are first extracted from images as their representations, and then the relevance/matching are evaluated using the representations so that the ranking can be conducted.
The two components of feature extraction and relevance/matching evaluation are of equal importance. This can be seen at the early stage of CBIR, when the same amount of attention was put on the two components \cite{smeulders2000content}.
However, most of the attention shifted to the feature extraction greatly as soon as the deep learning was introduced \cite{chen2021deep,zheng2017sift}. 

With deep learning, we can build complex models for image representations in parameter spaces of dimensionality up to billions \cite{kalantidis2016cross,babenko2015aggregating,arandjelovic2016netvlad}. This is a substantial improvement over the early methods in which much simpler models were hand-crafted in feature spaces with dimensionality of hundreds or several thousand \cite{sivic2003video,lowe2004distinctive,jegou2008hamming,jegou2010aggregating,perronnin2010improving}.
More importantly, deep features can be learned through gradient-based backprogation in an unified and automatic manner rather than through the dull feature engineering \cite{jegou2008hamming,zheng2014packing,arandjelovic2012three}.
Thanks to the richness and easy-to-learn nature of deep features, the feature representation learning is becoming the core of CBIR \cite{chen2021deep}, while the relevance/matching evaluation receives less and less attention, and has degenerated into elementary metrics (e.g., cosine similarity, Euclidean distance) in most of the current methods \cite{tolias2015particular,jimenez2017class,xu2018iterative}.

While promising results are being constantly reported, we argue that the emphasis on the superiority of representation learning over the relevance/matching evaluation may have taken us away from the spirit of retrieval.
The human logic to identify the relevant images is indeed based on the matching, in which the correspondence of global/local salient characteristics (e.g., colors, textures, shapes) between the query and targets acts as the core evidence. This is the reason that at the early stage of CBIR, the matching has been modeled explicitly (e.g., matching of local features extracted by SIFT/SURF used to be the most promising and popular scheme \cite{ke2004pca,arandjelovic2012three,philbin2010descriptor}).
It is not only the conceptual base of image retrieval but also the practical needs.
In many applications, it is useless to the users if the ranked lists are given alone without indicating the evidence. The most intuitive way to understand this is to think of the demand in text search domain, where it is important for the search engines to highlight the matched words in the returned results. Otherwise, the users would have to go through every returned document by themselves to verify if it is what they are looking for.
Similar demands exist in the image retrieval domain. For example, the indications of the locations/regions of the target protein/cell/lesion instances in the top-ranked medical images are generally expected. Otherwise, the technicians/doctors would have to put a great effort to verify each image before drawing conclusions on the disease \cite{kumar2013content,qayyum2017medical}.

However, contemporary deep retrieval methods have a limited ability to generate the explicit matching evidence.
Generally speaking, regardless of the models adopted directly with an off-the-shelf setting \cite{tolias2015particular,kalantidis2016cross,babenko2015aggregating,jimenez2017class} or indirectly through fine-tuning \cite{arandjelovic2016netvlad,chang2019explore,gordo2016deep}, the deep networks have originally been designed for classification tasks, where the 
goal is to learn high-level semantics instead of specific characteristics of instances. The \textit{domain gap} makes the characteristics (especially the local details) of the images hard to be backtracked or explained from the resulting representations, not to mention the correspondence between them.
From a computational point of view, the lack of explainability comes from the convolution that makes the local details diffuse to neighbour neurons when passing through layers. The more layers are stacked, the more serious the \textit{detail diffusion} is. This is what leads to the blackbox nature of deep networks. It can be seen from the Class Activation Mapping (CAM) results of the representations, which are commonly used as the evidence of inference but usually with arbitrary shapes. 
Therefore, there is a limited space left for the relevance/matching evaluation to collect evidence from the detail-diffused representations.
Efforts have been made for the region-specific representation generation in a number of methods such as: R-MAC \cite{tolias2015particular} that pools the feature maps by regions; Region Proposal Networks (RPNs) \cite{gordo2016deep} and Faster R-CNN \cite{salvador2016faster} for selective region feature extractions; multiple pass learning that feeds patches instead of the whole image into networks \cite{liu2015deepindex}; and Gaussian centered prior \cite{babenko2015aggregating}, Crow \cite{kalantidis2016cross}, and CAM+Crow \cite{jimenez2017class} that use attention mechanisms to highlight salient regions. 
Nonetheless, these efforts are just complementary, and the matching evidence generation has been neither set as the goal nor formulated explicitly. It is thus still difficult for these methods to generate matching evidence indications.

\begin{figure}
\label{fig:proposed}
    \centering
    \includegraphics[width=0.95\columnwidth]{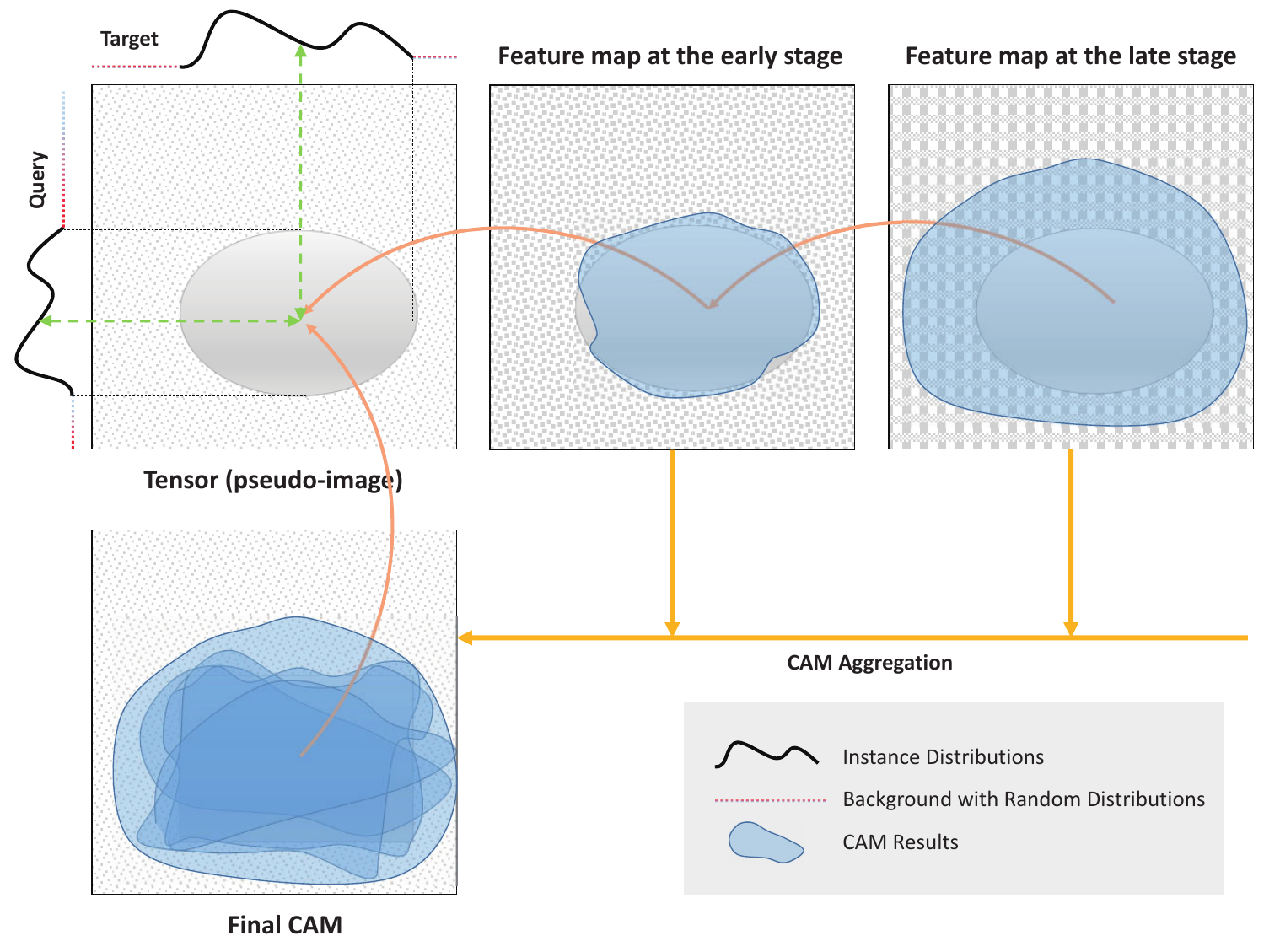}
    \caption{Illustration of the indicative image retrieval: the correspondence between the query and target images are encapsulated into a tensor for modeling the matching; the tensor is used as a pseudo-image and fed into deep networks with the relevance label of the image pair; the CAM results of the early and late layers are finally aggregated and decoded into evidence (matched patch pairs) of the prediction. The matched instance distributions form a densely dispersed hypersphere which is with a regular shape and thus more robust to the detail diffusion.}
    \label{fig:my_label}
\end{figure}

In this paper, we present a pilot study for reclaiming the important role of relevance/matching in deep learning era. 
To remove the effect of feature extraction, we have skipped the representation learning and started from modeling the matching directly. 
As none of the pre-trained models are employed, the \textit{domain gap} is therefore eliminated.
As illustrated in Figure~\ref{fig:proposed}, the matching modeling is implemented by first calculating the correlation of image patches using the pixels (without any feature extraction), and then encapsulating the results into a multi-view 4D tensor as the pseudo-image for deep networks.
In the pseudo-image, a pair of matched instance distributions will form a high-correlation pattern which densely disperses over a hypersphere in the tensor.
As the hypersphere has a much regular shape than the arbitrary ones of the instances in the original images, the \textit{detail diffusion} will affect more on its range, while the shape will be better reserved.
Therefore, we can reconstruct the shape from the aggregated CAM result if it has been frequently adopted at different layers for inference.
%
%We do not rely on the accuracy of the reconstructed shape too much. As long as the dominating tensor region has been located. 
%
The matching evidence indications can then be generated by decoding from the reconstructed hypersphere, in which the correspondence of patch pairs has been explicitly ``packed'' at the beginning. 
%
% The method has been illustrated in Figure~\ref{fig:proposed}, which has advantages over previous deep image retrieval models as follows.
% \begin{itemize}
%     \item \textbf{Free of Representation Learning Problems}: It is obvious that the framework has prevented the problems from happening by skipping feature extraction. It is thus free of most of the representation learning problems such as the domain gap between the classification and retrieval \cite{tolias2015particular,kalantidis2016cross,babenko2015aggregating}, detail diffusion \cite{yang2017two,ozaki2019large}, and the balance between global and local feature embedding \cite{gong2014multi,yue2015exploiting,cao2020unifying}.
    
%     \item \textbf{Deep Architecture Adaptability}: The framework has turned the similarity-based evaluation into a binary classification problem. This is what most of the deep networks has been designed for, and the power of deep learning can thus be better leveraged in this case. %query dependent
    
%     \item \textbf{Evidence Generation}: As shown in Figure~X, the correspondence of patch pairs have been ``packed'' into the input tensor, what diffuses during the convolution is the ``relation'' of patches rather than the patches themselves (no matter how many layers they have been passed in the network). This give us opportunity to backtrack the patches that result in the prediction of relevance. Therefore, the matching evidence indications can be generated in an explicit way. More details can be found in Section~Z.
    
%     \item \textbf{Retrieval Tractability}:
% \end{itemize}

The contributions of this paper include: 1) we demonstrate that superior retrieval performance can be obtained by modeling the matching directly without depending on representation learning and pre-trained models; and 2) we address the blackbox nature of the deep networks in the image retrieval scenario, and increase the explainability by generating matching evidence indications.

%%%%%%%%%%%%%%%%%%%%%%%%%%%%%%%%%%%%%%%%%%%%

\section{Related Work}
\subsection{A Brief for Image Retrieval}
Image retrieval can date back to 1970s when it was considered as a task of searching images on their text annotations. It had been a commonly adopted method for visual content management until the early 1990s when the retrieval community started to model it as a matching problem based on the visual features extracted from the images directly \cite{jain1993nsf}.
Global features (e.g., colors, textures), which describe images as a whole, were prevalent clues since then \cite{smeulders2000content}. However, the research attention has soon been shifted to the local features, largely because of the success of SIFT \cite{lowe2004distinctive} which creates more robust representations (than the global features) for addressing the variance of image changes (e.g., illumination, scale, translation).
More importantly, by considering the local features as the ``visual'' words for images, it makes the image retrieval easy to conduct by borrowing techniques from the text search domain which was more extensively studied at that time. The Bag-of-Words (BoW) \cite{sivic2003video} is such a model which encodes visual words in an image into a compact and representative feature vector for more efficient retrieval. The configuration of SIFT+BoW has been a prominent solution for over a decade \cite{zheng2017sift}.

It was not until the early 2010s that the deep learning stands to take over the position. In 2012, Krizhevsky \textit{et al.} \cite{krizhevsky2012imagenet} has demonstrated the superior performance of AlexNet over previous methods in classification tasks of ILSRVC 2012.
The convolutional neural networks (CNN) shortly gained popularity in almost all types of computer vision tasks including image retrieval, because of its aforementioned representativeness and easy-to-learn nature (over hand-crafted representations).

\subsection{Deep Feature Extraction}
The representations can be obtained by employing the pre-trained models (e.g., VGG \cite{simonyan2014very}, GoogleNet \cite{szegedy2015going}, ResNet \cite{he2016deep}) for classification tasks either in an off-the-shell manner \cite{tolias2015particular,kalantidis2016cross,babenko2015aggregating,jimenez2017class} or fine-tuned scheme \cite{arandjelovic2016netvlad,chang2019explore,gordo2016deep}.
A feature vector is usually generated from the activation of the full-connected layers (FCs) for global clues \cite{babenko2014neural,cao2016quartet}, or composed from the feature maps of convolutional layers for local patterns \cite{tolias2015particular,kalantidis2016cross,babenko2015aggregating}. However, this is just a principle in a general sense. It is a popular topic to find which layers (or fusion of them) generate the best representations for different tasks/datasets \cite{yu2017exploiting,li2016exploiting,yu2018multi}.
Moreover, feature aggregation such as sum/average, max pooling, and embedding (e.g., BoW \cite{sivic2003video}, VLAD \cite{jegou2010aggregating}, FV \cite{perronnin2010improving}) is also applied to improve the discriminativeness and compactness of the representations.
In fact, representation learning is still an ongoing topic, and numerous methods are proposed in recent years which are conducted either supervised \cite{arandjelovic2016netvlad,babenko2014neural,gordo2016deep} or non-supervised \cite{radenovic2018fine,xu2018iterative}, and directly concatenated \cite{babenko2015aggregating,razavian2016visual} or hash embedded \cite{huang2018object,liu2016deep}.
We will skip these methods here because we have skipped representation learning in our framework. The reader can refer to \cite{chen2021deep,zheng2017sift} for more details.

\subsection{Verification-based Learning}
Although the feature extraction has received the overwhelming attention in retrieval community, there are still verification-based networks (VBNs) which have put the focus on the matching (e.g., Simo  et al. \cite{simo2015discriminative}, Ong et al. \cite{ong2017siamese}, Zhou et al. \cite{zheng2019hardness}). 
Instead of employing the simple similarity metrics, VBNs use metric learning to find an optimal measure to distinguish relevant pairs of images from the irrelevant ones. It is often implemented with Siamese (triplet) networks \cite{simo2015discriminative,ong2017siamese,arandjelovic2016netvlad} in which pairs (triplet sets) of images are fed into a two-stream (three-stream) network to learn how to classify the labels (i.e., relevant vs. irrelevant).
While promising results being obtained, VBNs rely heavily on the basic Siamese (triple) networks, which have not been studied to the same extent as CNNs. The design of the structures and loss functions thus raise further challenges \cite{chen2021deep}.
Nonetheless, NBNs are also using the CNNs as the backbone networks for feature extractions, which means that the metrics are learnt from the detail-diffused deep representations, and the results of matching are still not easy to be backtracked into regions of the original images.
It is worth mentioning that some effort have indeed been made on generating region-aware VBNs using region proposal networks (RPNs) \cite{gordo2016deep} or Faster R-CNN \cite{salvador2016faster}.
Nevertheless, the resulting vectors are fused or aggregated later in the networks, because they are not designed for explicit matching.

While our focus in this paper is to build explicit matching rather than metrics, the learning scheme we have adopted shares certain similarities with VBNs, in the way the images are fed as pairs and the learning has used relevance labels.
However, there are essential differences from VBNs. First of all, 
the purpose of learning is to generate matching evidence indications not the metrics (although the results can be used as metrics as well).
Secondly, we are using the explicit matching information to compose pseudo images (i.e., multi-view tensors) as the input, which allows the backtracking of matched regions.
Thirdly, the input is query-dependent rather than the fixed feature vectors extracted by pre-trained CNNs. This is closer to the spirit of retrieval \cite{croft1981incorporating,bartell1994automatic,lee1995combining}.
Finally, we use the sophisticated and end-to-end CNNs for the learning rather than pre-matured CNNs+multi-stream networks. The learning is much more straightforward than multi-stream networks, thus efficacy is better guaranteed. 
More details will be given in Section~\ref{sec:method}.

\subsection{Region-Specific Attention Mechanisms}
In terms of using the attention mechanisms to highlight the specific regions in retrieval, non-parametric methods (e.g., Gaussian center prior \cite{babenko2015aggregating} and CroW \cite{kalantidis2016cross}) and parametric methods (e.g., \cite{mohedano2018saliency,yang2017two}) have also been proposed.
Our method is more like a combination of these two types of methods, in the sense that we use the hypersphere with Gaussian density as a prior attention but leave all parameters to be learnt through the standard backpropagation.
Nonetheless, our attention mechanism is essentially different from these methods. The regions are highlighted in previous methods because of their salience in the original images. It is determined by the static characteristics of the images. In our method, the attention is given to the most salient matching (i.e., correspondence of patches). It is dynamically determined in the way that different regions will be highlighted based on the query.

The proposed method is also related to several tensor-as-pseudo-image work (e.g., the 4D tensors \cite{rocco2020ncnet,rocco2017convolutional}, the collocative tensors \cite{wei2021deep}).
However, the tensors of these methods are built on deep features for solving problems in domains different from image retrieval (e.g., 3D reconstruction \cite{rocco2020ncnet}, geometry matching \cite{rocco2017convolutional}, multi-label medical image recognition \cite{wei2021deep}). The motivations and implementations are all different as well.

% In terms of the learning framework, we are also inspired by the collocative learning prototype , in which the authors argue that traditional CNNs are in fact learning the unary mapping (an image to a label), and are not feasible to learn the a binary mapping (a Immunofixation Electrophoresis band pair to a label).
% %
% Therefore, the collocative learning has been proposed to transform the binary relation into a pseudo-image so that the CNNs can be used.
% %
% We have followed this thread because the matching of patches are also a high-order relation (i.e., quaternary relation indeed). However, the framework has been implemented in a different way to adapt the challenges raised in the new problem domain of image retrieval. 

%%%%%%%%%%%%%%%%%%%%%%%%%%%%%%%%%%%%%%%%%%%%

\section{Indicative Image Retrieval}
\label{sec:method}
In this section, we first introduce the base-frame of the indicative image retrieval, namely, the modeling of matching evidence and the backtracking of the evidence for indication generation.
Then we show that by modeling the matching explicitly, the framework generates new space for exploring matching-oriented techniques, including the region-specific search through hypersphere attention regulation, and the indicative pseudo-relevance feedback. 

\subsection{Modeling the Matching Evidence}
To bypass the aforementioned issues of representation learning, we start from modeling the matching (correspondence) of image patches directly. 
It is easy to decompose an image into patches using the image patch generation schemes (e.g., rigid grid \cite{cao2017local}, spatial pyramid modeling \cite{liu2015deepindex}, dense patch sampling \cite{sharif2014cnn}, region proposals \cite{gordo2016deep}). The inter-patch correlation can then be encapsulated into a 4D tensor as the pseudo-image. 

Let us assume an image is decomposed by a grid of $m$ rows and $n$ columns, the correlation of a pair of patches $\{p_{ij},q_{kl}\}$ is evaluated with a similarity/distance metric $\phi(p_{ij},q_{kl})$, where $0\leq i,k\leq m$ and $0\leq j,l\leq n$ are the indices of the patches $p$ and $q$ from the target and query image, respectively.
The tensor is denoted as 
\begin{equation}
    \boldsymbol{R}_{\phi}(i,j,k,l)=\phi(p_{ij},q_{kl}), \boldsymbol{R}_{\phi}\in \mathbb{R}^{m\times n\times m\times n}.
\end{equation}
By varying the $\phi$ from a metric set of $M=$\{\textit{Cosine Similarity, Euclidean distance, Manhattan Distance, Mahalanobis Distance, ...}\}, a multi-view tensor $\boldsymbol{\mathcal {T}}$ of order-5 can then be composed by the concatenation of these $\boldsymbol{R}$'s as
\begin{equation}\label{eq:concat_single_view}
	\boldsymbol{\mathcal {T}}=\{\boldsymbol{R}_{\phi_1},\boldsymbol{R}_{\phi_2},\dots,\boldsymbol{R}_{\phi_{|M|}}\}=\bigcup_{i=1}^{|M|}\boldsymbol{R}_{\phi_i}.
\end{equation}
The tensor can then be used as a pseudo-image for the input of any sophisticated CNNs (VGG \cite{simonyan2014very}, GoogleNet \cite{szegedy2015going}, ResNet \cite{he2016deep} among others). 
In this case, we can leverage the power of sophisticated deep learning to complete the inference of the relevance.
Note that the $\boldsymbol{R}$'s are indeed imitating the channels of the pseudo-image, and the 
first 4 dimensions of the tensor $\boldsymbol{\mathcal {T}}$ are imitations of the spatial dimensions.

\subsection{Generation of Evidence Indications}
\label{sec:evidence_generation}
Since the matching evidence is explicitly ``packed'' into the pseudo-image which is more robust to detail-diffusion, the evidence can thus be backtracked by decoding the Class Attention Mapping (CAM) results for the original matching pairs and evidence indication generation.

\subsubsection{CAM Aggregation}
Let us collect the inference clues by aggregating CAM results of all layers into a final CAM first.
Denote the set of all layers as $\{\gamma\}$, in which each layer $\gamma$ has a feature map $\boldsymbol{\mathcal{F}^\gamma}$. A CAM (saliency map) $\boldsymbol{\mathcal{M}^\gamma}$ for layer $\gamma$ can be generated from $\boldsymbol{\mathcal{F}^\gamma}$ by using Grad-CAM \cite{selvaraju2017grad}.
Note that each $\boldsymbol{\mathcal{M}^\gamma}$ is a 4D tensor which can be re-scaled into the same size as that of the spatial dimensions of $\boldsymbol{\mathcal {T}}$.
The final CAM is then generated from all $\boldsymbol{\mathcal{M}^\gamma}$'s using average pooling as
\begin{equation}
	\boldsymbol{\mathcal{M}^*}=\frac{1}{|\{\gamma\}|}\sum_{\gamma\in\{\gamma\}}\boldsymbol{\mathcal{M}^\gamma},\hspace{0.1in} \boldsymbol{\mathcal{M}^*}\in \mathbb{R}^{m\times n\times m\times n}.
\end{equation}

\subsubsection{CAM Decoding}
Each element of the final CAM (denoted as $\boldsymbol{\mathcal{M}^*}_{ijkl}$) indeed gives a rating of how much a patch pair $\{p_{ij},q_{kl}\}$ has contributed for predicting the relevance label.
Therefore, we can assign this contribution score to each of its member patches $p_{ij}$ and $q_{kl}$, respectively. Once all patches have collected their scores from its host pairs, it works like a voting scheme in which the contributions of patches are evaluated.
This can be done by contracting the $\boldsymbol{\mathcal{M}^*}$ into its first two and last two dimensions, respectively, as
\begin{equation}
\begin{split}
    \boldsymbol{\mathcal{P}}(i,j)=&\sum_{k,l}\boldsymbol{\mathcal{M}^*}_{ijkl}{\bullet}_{3,4}^{1,2}\mathbbm{1}^{m\times n}_{kl},\\
    \boldsymbol{\mathcal{Q}}(k,l)=&\sum_{i,j}\mathbbm{1}^{m\times n}_{kl}\bullet_{1,2}^{1,2}\boldsymbol{\mathcal{M}^*}_{ijkl},
\end{split}
\end{equation}
where $\mathbbm{1}^{m\times n}_{kl}$ is a ones tensor, $A\bullet_{a,b}^{c,d}B$ is a double contraction operator that represents the inner product of the $(a,b)^{th}$ mode of $A$ with the $(c,d)^{th}$ mode of $B$, $\boldsymbol{\mathcal{P}}$ (or $\boldsymbol{\mathcal{Q}}$) is a 2D matrix in which the $(i,j)^{th}$ ($(k,l)^{th}$) element indicates the contribution score of the target (or query) patch $p_{ij}$ ($q_{kl}$).

\subsection{Hypersphere Attention Regulation}
In Figure~\ref{fig:proposed}, we show that the matched instance distributions form a hypersphere pattern within the input tensor. It is of more regular shapes than the arbitrary ones of the instances in the original images, which is also easier to be detected and more robust to \textit{detail diffusion}. 
In this section, we further take advantage of this characteristics to guide the verification process to focus on the principal matching evidence with a hypersphere attention mask (regulator).

The idea is to build a parameterized hypersphere attention mask $\boldsymbol{\Theta}$ to regress the shape and position of the evidence pattern. In case that this attention mask matches the evidence pattern with a similar shape and position (learnt through regression), the maximum ``resonance'' will be created which allows more information from the evidence pattern to pass through the mask (while depress the passage of the rest of the regions).
We can apply it to every layer $\{\gamma\}$ to modify its feature map $\boldsymbol{\mathcal{F}^\gamma}$ as
\begin{equation}\label{eq:regulation}
	\boldsymbol{\hat{\mathcal{F}}}^\gamma=\boldsymbol{\mathcal{F}}^\gamma\odot \boldsymbol{\Theta}^\gamma + \boldsymbol{\mathcal{F}}^\gamma, \gamma\in\{\gamma\}
\end{equation}
where $\boldsymbol{\hat{\mathcal{F}}}^\gamma$ is the updated feature map, $\odot$ is the Hadamard product, and the $\boldsymbol{\Theta}^\gamma$ is the mask at the layer $\gamma$.
The mask is indeed also a shape regulator. The effect of the \textit{detail diffusion} will be reduced because $\boldsymbol{\Theta}^\gamma$'s will work collaboratively to help the evidence pattern keep its shape as a hypersphere.

The parameterized mask $\boldsymbol{\Theta}^\gamma$ can be learnt through a function $\omega(i,j,k,l|\vec{c},N(\mu,\sigma))\in[0,1]: \mathbb{Z}^{+4\times 1}\to \mathbb{R}$ which assigns for the $(i,j,k,l)^{th}$ element of the feature map $\boldsymbol{\mathcal{F}^\gamma}$ a weight of $\boldsymbol{\Theta}^\gamma_{ijkl}=\omega(i,j,k,l)$ determined by the center of the mask $\vec{c}\in \mathbb{Z}^{+4\times 1}$ (to regress the location of the evidence hypersphere), and a Gaussian distribution $N(\mu,\sigma)$ 
over the distance $d$ of the $(i,j,k,l)^{th}$ element to the center (to regress the density distribution of the evidence hypersphere). It is written as
\begin{equation}\label{eq:har}
	\begin{split}
		\boldsymbol{\Theta}^\gamma_{ijkl}&=\omega(i,j,k,l|\vec{c},N(\mu,\sigma))\\
		&=\frac{1}{\sqrt{2\pi}\sigma}\exp{\left(-\frac{(d-\mu)^2}{2\sigma^2}\right)},\\
		\vec{x}&=[i,j,k,l]^\top,\\
		d&=\|\vec{x}-\vec{c}\|^2,
			%=&\underbrace{\alpha}_{\text{scalar}}\cdot\boldsymbol{\cos}\left(2\pi T\underbrace{\frac{|i-j|}{m}}_{\text{normalizer}}+\underbrace{\beta}_{\text{offset}}\right)+\underbrace{\gamma}_{\text{adjuster}}   
	\end{split}
\end{equation}
where $\vec{x}$ is the coordinate vector of the $(i,j,k,l)^{th}$ element. We learn the $6$ parameters (i.e., 4 for $\vec{c}$, 2 for $N(\mu,\sigma)$) in Eq.~(\ref{eq:har}) through the backpropagation by replacing the CNN functions for updating the gradients as
\begin{equation}
	\begin{split}
		\frac{\partial Y^c}{\partial \mu}&=\sum_{i,j,k,l}\frac{\partial Y^c}{\partial \boldsymbol{\Theta}^\gamma_{ijkl}}\frac{\partial \boldsymbol{\Theta}^\gamma_{ijkl}}{\partial \mu}\\
		&=\sum_{i,j,k,l}\frac{\partial Y^c}{\partial \boldsymbol{\Theta}^\gamma_{ijkl}}\Bigg[\frac{\partial \frac{1}{\sqrt{2\pi}\sigma}\exp{\left(-\textstyle\frac{(d-\mu)^2}{2\sigma^2}\right)}}{\partial \mu}\Bigg]\\
		&=\sum_{i,j,k,l}\frac{\partial Y^c}{\partial \boldsymbol{\Theta}^\gamma_{ijkl}}\Bigg[\frac{1}{\sqrt{2\pi}\sigma}\exp{\left(-\textstyle\frac{(d-\mu)^2}{2\sigma^2}\right)}\left(\textstyle\frac{d-\mu}{\sigma^2}\right)\Bigg]\\
		&=\sum_{i,j,k,l}\frac{\partial Y^c}{\partial \boldsymbol{\Theta}^\gamma_{ijkl}}\frac{d-\mu}{\sigma^2}\boldsymbol{\Theta}^\gamma_{ijkl}\\
		&=\frac{1}{\sigma^2}\sum_{i,j,k,l}\frac{\partial Y^c}{\partial \boldsymbol{\Theta}^\gamma_{ijkl}}\left(d-\mu\right)\boldsymbol{\Theta}^\gamma_{ijkl},\\
	\end{split}
\end{equation}

%\vspace{-3cm}

\begin{equation}
	\begin{split}
		\frac{\partial Y^c}{\partial \sigma}&=\sum_{i,j,k,l}\frac{\partial Y^c}{\partial \boldsymbol{\Theta}^\gamma_{ijkl}}\frac{\partial \boldsymbol{\Theta}^\gamma_{ijkl}}{\partial \sigma}\\
		&=\sum_{i,j,k,l}\frac{\partial Y^c}{\partial \boldsymbol{\Theta}^\gamma_{ijkl}}\Bigg[\frac{\partial \frac{1}{\sqrt{2\pi}\sigma}\exp{\left(-\textstyle\frac{(d-\mu)^2}{2\sigma^2}\right)}}{\partial \sigma}\Bigg]\hspace{0.47in}\\
		&=\sum_{i,j,k,l}\frac{\partial Y^c}{\partial \boldsymbol{\Theta}^\gamma_{ijkl}}\Bigg[-\frac{1}{\sqrt{2\pi}\sigma^2}\exp{\left(-\textstyle\frac{(d-\mu)^2}{2\sigma^2}\right)}+\\
		&\hspace{0.4in}\frac{1}{\sqrt{2\pi}\sigma}\exp{\left(-\textstyle\frac{(d-\mu)^2}{2\sigma^2}\right)}\frac{(d-u)^2}{\sigma^3}\Bigg]\\
		&=\sum_{i,j,k,l}\frac{\partial Y^c}{\partial \boldsymbol{\Theta}^\gamma_{ijkl}}\left(-\frac{1}{\sigma}\boldsymbol{\Theta}^\gamma_{ijkl}+\frac{(d-\mu)^2}{\sigma^3}\boldsymbol{\Theta}^\gamma_{ijkl}\right)\\
		&=\frac{1}{\sigma^3}\sum_{i,j,k,l}\frac{\partial Y^c}{\partial \boldsymbol{\Theta}^\gamma_{ijkl}}\left((d-\mu)^2-\sigma^2\right)\boldsymbol{\Theta}^\gamma_{ijkl},\\
	\end{split}
\end{equation}
\vspace{-1.3cm}

\begin{equation}
	\begin{split}
		\frac{\partial Y^c}{\partial \vec{c}}&=\sum_{i,j,k,l}\frac{\partial Y^c}{\partial \boldsymbol{\Theta}^\gamma_{ijkl}}\frac{\partial \boldsymbol{\Theta}^\gamma_{ijkl}}{\partial d}\frac{\partial d}{\partial \vec{c}}\\
		&=\sum_{i,j,k,l}\frac{\partial Y^c}{\partial \boldsymbol{\Theta}^\gamma_{ijkl}}\frac{\partial \frac{1}{\sqrt{2\pi}\sigma}\exp{\left(-\textstyle\frac{(d-\mu)^2}{2\sigma^2}\right)}}{\partial d}\frac{\partial \|\vec{x}-\vec{c}\|^2}{\partial \vec{c}}\\
		&=\sum_{i,j,k,l}\frac{\partial Y^c}{\partial \boldsymbol{\Theta}^\gamma_{ijkl}}\Bigg[\frac{1}{\sqrt{2\pi}\sigma}\exp{\left(-\textstyle\frac{(d-\mu)^2}{2\sigma^2}\right)}\\
		&\hspace{0.4in}\left(-\frac{2(d-\mu)}{2\sigma^2}\right)\left(-2\|\vec{x}-\vec{c}\|\right)\Bigg]\\
		&=\sum_{i,j,k,l}\frac{\partial Y^c}{\partial \boldsymbol{\Theta}^\gamma_{ijkl}}\left(\boldsymbol{\Theta}^\gamma_{ijkl}\frac{2(d-\mu)\|\vec{x}-\vec{c}\|}{\sigma^2}\right)\\
		&=\frac{2}{\sigma^2}\sum_{i,j,k,l}\frac{\partial Y^c}{\partial \boldsymbol{\Theta}^\gamma_{ijkl}}\left((d-\mu)\|\vec{x}-\vec{c}\|\right)\boldsymbol{\Theta}^\gamma_{ijkl},\\
	\end{split}
\end{equation}

\vspace{-0.5cm}

\subsection{Indicative Pseudo Relevance Feedback}
Pseudo-relevance feedback (PRF) has been a popular re-ranking method for image retrieval since the late 2000s \cite{zhou2003relevance}. It pools the top-ranked results in the initial round for more informative clues about the query.
The clues are used to refine the query (e.g., through query expansion) or retrain the model (e.g., a classifier which distinguishes relevant from irrelevant ones). A new round of search can then be conducted for improving the performance.
Its effectiveness has been validated in many applications. However, it receives less attention recently. One reason is that the query refinement is usually conducted through a prior which requires an explicit understanding of the query vector. The deep features are thus less feasible because of their blackbox nature.
In this section, we show PRF can also be integrated into indicate image retrieval with the explicit query refinement.

Following the assumption of PRF, we assume the top-$n$ results in the initial ranked list are relevant target images. Each of them is associated with two contribution maps of $\boldsymbol{\mathcal{P}}$ and $\boldsymbol{\mathcal{Q}}$ for the target and query images, respectively, which have been decoded from the matching evidence indications using method introduce in Section~\ref{sec:evidence_generation}. In $\boldsymbol{\mathcal{Q}}$, the regions better matched than others have been indicated. Therefore, we can summarize the $n$ $\boldsymbol{\mathcal{Q}}$'s into an optimize one $\boldsymbol{\mathcal{Q}}^*$ so that the regions which are more informative (i.e., more consistently matched across the $n$ targets) will be found. Then we can refine the query by using $\boldsymbol{\mathcal{Q}}^*$ as a weighting mask.
Among a lot of options to summarize the $\boldsymbol{\mathcal{Q}}$'s, we use the simplest ones of \textit{average} and \textit{max} pooling, because the purpose is to explore the PRF in indicative retrieval framework rather than building a sophisticated PRF strategy.
The pooling operations are written as
\begin{equation}
    \boldsymbol{\bar{\mathcal{Q}}^*}=\frac{1}{n}\sum_{i=1}^n \boldsymbol{\mathcal{Q}}_i,\hspace{0.1in}\boldsymbol{\hat{\mathcal{Q}}^*}=\max_{1\leq i\leq n} \boldsymbol{\mathcal{Q}}_i,
\end{equation}
where the $\boldsymbol{\bar{\mathcal{Q}}^*}$ and $\boldsymbol{\hat{\mathcal{Q}}^*}$ are the optimized contribution maps (masks) using the \textit{average} and \textit{max} pooling, respectively. Both pooling operations are conducted element-wisely.
%
%The $\boldsymbol{\mathcal{Q}}^*_{avg}$ can be used for improving precision while $\boldsymbol{\mathcal{Q}}^*_{max}$ for recall.

\begin{figure}[t]
	\centering
	\includegraphics[width=0.48\textwidth]{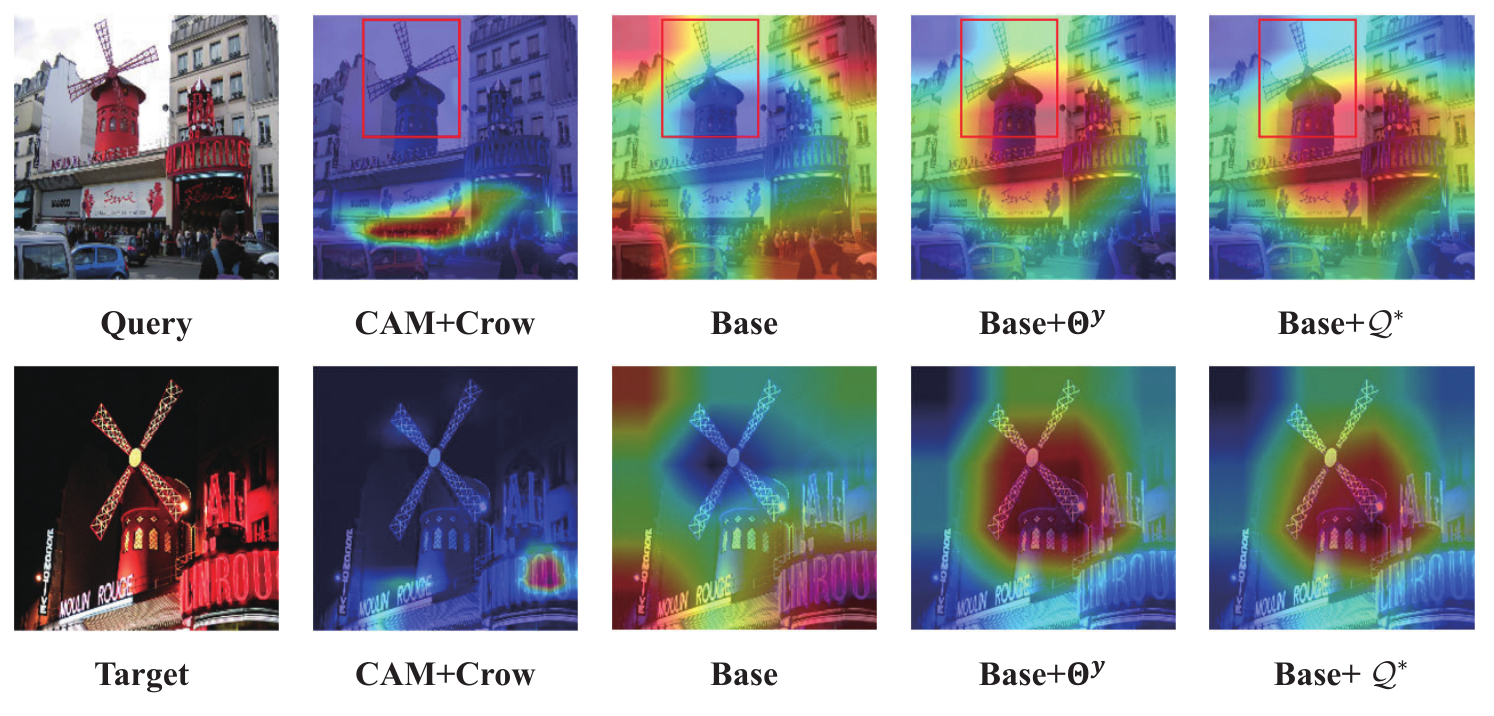}
	%\vspace{-0.1in}
	\caption{The CAM Maps of the CAM+Crow and the Decoded Evidence Maps of our Indicative Retrieval Models. The results of CAM+Crow show no matched regions between the query and target, while the matched regions by the indicative retrieval are evident .
	The effectiveness of the hypersphere attention regulation (HAR) and the pseudo-relevance feedback (PRF) can also be observed from the refined maps with better attention on the Moulin Rouge than those of the Base model (with the 4-D tensor only) which collect evidence all over the images.}\label{fig:samples}
	%\vspace{-0.2in}
\end{figure}

An example of the contribution maps is shown in Figure~\ref{fig:samples}. In contrast to those of CAM+Crow \cite{jimenez2017class} (the region-specific attention method with recognized performance in literature), our methods show better attention on the Moulin Rouge and more reasonable matched regions between the query and target. %
This is the advantage that the query and target are considered as a pair and used as co-reference, rather than being treated separately in traditional methods like CAM+Crow.
Another interesting thing in Figure~\ref{fig:samples} is that in the results by the Base (with 4-D tensor only), attention is ``inverted'' which means that the matched regions of the surrounding background are used as the evidence for inference and those from the Moulin Rouge are excluded. It is an indication that an ``inverse'' hypersphere may have been created in the conventional layers to support the prediction.
Predicting based on the background matching is indeed not wrong, because the model apparently learns to answer the question of ``what it is'' with the evidence of ``what is not''.   
However, this is not instance-oriented. Fortunately, this can be corrected with either the hypersphere attention regulation (HAR) or pseudo-relevance feedback (PRF).

%%%%%%%%%%%%%%%%%%%%%%%%%%%%%%%%%%%%%%%%%%%%

\begin{table*}
	\centering \caption{Performance comparison to baselines. The hypersphere attention regulation (HAR) is denoted as $\boldsymbol{\Theta}^\gamma$. The pseudo-relevant feedback (PRF) using the \textit{average} and \textit{max} pooling are denoted as $\boldsymbol{\bar{\mathcal{Q}}}^*$ and $\boldsymbol{\hat{\mathcal{Q}}}^*$, respectively. The mAP@$k$ denotes the mAP evaluated through the top-$k$ returned results. The mIoU is the mean of IoUs over thresholds over [0.0 0.9]. The best results are in bold font.} \label{tb:map}
	\setlength\extrarowheight{1.5pt}\setlength\tabcolsep{2pt}
	\renewcommand{\multirowsetup}{\centering}
	{\footnotesize%begin table fontsize
	\begin{tabular}{l|ccccc|ccccc}
		\hline
		 & \multicolumn{5}{c|}{\small Oxford-5k} &\multicolumn{5}{c}{\small  Paris-6k}\\ 
		 \cline{2-11}
		{\small Method}   & mAP & mAP@5  &mAP@10 &mAP@20  & mIoU  & mAP & mAP@5  &mAP@10 &mAP@20 & mIoU \\ \hline
			CAM+Crow &82.53$\pm$4.02  &95.43$\pm$3.79        &91.53$\pm$2.92      &88.49$\pm$2.82              &11.95$\pm$8.26 
			&71.38$\pm$4.83 & 96.54$\pm$3.75          & 93.61$\pm$3.70      &89.73$\pm$5.02            &9.51$\pm$9.33 \\
			Base &97.71$\pm$2.12               		    & \textbf{100.00$\pm$0.00}    &\textbf{99.98$\pm$0.06} &\textbf{99.62$\pm$0.60}		&21.49$\pm$11.37  
			& 94.05$\pm$7.22       	        & \textbf{100.00$\pm$0.00}    & 99.78$\pm$0.44          &99.53$\pm$0.69				&18.66$\pm$9.59\\
			Base+$\boldsymbol{\Theta}^\gamma$ &\textbf{97.77$\pm$2.12}           &\textbf{100.00$\pm$0.00}      &99.70$\pm$0.91     &99.49$\pm$1.47            &22.51$\pm$10.66 
			& 97.43$\pm$0.98                   & 99.56$\pm$0.89               &99.53$\pm$0.94          &99.49$\pm$1.01                &18.66$\pm$9.59\\
			Base+$\boldsymbol{\Theta}^\gamma$+$\boldsymbol{\bar{\mathcal{Q}}}^*$ &96.02$\pm$2.49                 &99.18$\pm$1.77    &98.14$\pm$2.26       &97.60$\pm$2.33           &\textbf{22.84$\pm$10.86}  
			& \textbf{97.81$\pm$0.90}           &\textbf{100.00$\pm$0.00}         &\textbf{99.95$\pm$0.10}          &\textbf{99.80$\pm$0.32}           &19.57$\pm$10.17\\
			Base+$\boldsymbol{\Theta}^\gamma$+$\boldsymbol{\hat{\mathcal{Q}}}^*$ &94.83$\pm$2.41                  &97.57$\pm$3.33      &97.06$\pm$2.97      &96.71$\pm$2.54           &22.54$\pm$11.16  
			&96.90$\pm$1.11                     &\textbf{100.00$\pm$0.00}        &99.85$\pm$0.31          &99.65$\pm$0.43            &\textbf{20.00$\pm$10.38}\\
			\hline
	\end{tabular}
    }%set table fontsize
\end{table*}

\section{Experiments}
%\subsection{Setup}
We have employed the Oxford-5K \cite{philbin2007object} ($5,062$ images with 55 queries) and Paris-6k \cite{philbin2008lost} ($6,412$ images with 500 queries) for evaluation. These two datasets are considered as the most popular ones for image retrieval in literature. Although Filipe \textit{et al}. \cite{radenovic2018revisiting} updated the annotations and protocols of these datasets recently, we are using the original versions to be consistent with most of the previous methods.
We use the rigid-grid of $14$-by-$14$ to decompose the images into patches.

The mean Average Precision (mAP) is adopted for evaluating the retrieval performance.
The Intersection over Union (IoU) \cite{chattopadhay2018grad} is used to evaluate the accuracy of the evidence indications based on the ratio of the overlap between the threshold-ed $\boldsymbol{\mathcal{P}}$ (contribution map of the target image) and the Regions of Interest (RoI). For traditional methods, we evaluate IoU on their CAM results if the models are available.
We have annotated the RoIs of images in Oxford-5K and Paris-6k. The annotations and the source code are available at http://(open upon acceptance).
All performance are reported with 10-fold cross validations.

\subsection{Baselines}
We select two baselines as follows:
\begin{itemize}
    \item CAM+Crow \cite{jimenez2017class}: Our methods focus on giving matching evidence indications. However, only the region-specific methods take this into consideration using attention mechanisms, and the matching between regions are usually used to study the effectiveness of these methods. Therefore, we select CAM+Crow as the representative, because it is reported with the best performance among region-specific methods in literature \cite{chen2021deep}.
    
    \item Indicative Retrieval (Base): Our method is compatible to all CNNs. Among popular ones, we employ ResNet18 \cite{he2016deep} for its simpleness and efficiency. This makes it suitable for large-scale ablation study. The multi-view tensor $\boldsymbol{\mathcal {T}}$ is constructed based on patches obtained with the $14$-by-$14$ grid, the evidence indications are generated with the approach introduced in Section~\ref{sec:evidence_generation}. We denote this run as Base hereafter.
\end{itemize}
The baseline results of mAP are shown in Table~\ref{tb:map}. The Base outperforms the CAM+Crow by $18.39\%$ and $31.76\%$ on Oxford-5k and Paris-6k, respectively. This is a supervising result, because CAM+Crow is employing VGG16 as its backbone, whose performance has been validated in a lot of applications.
The backbone is also with $12.55$ times more parameters than those of ResNet18 (138M vs. 11M), and is pre-trained on large-scale dataset of ImageNet \cite{russakovsky2015imagenet}, while the Base does not employ any pre-training.
The comparison of these two baselines on IoUs (over thresholds) can be found in Figure~\ref{fig:iou}. 
It is not surprising that the Base outperforms the CAM+Crow by $209.19\%\pm 146.96$, because the latter is not designed for indicative region matching.

\subsection{Hypersphere Attention Regulation}
In Table~\ref{tb:map}, by including hypersphere attention regulation, the Base+HAR outperforms CAM+Crow by $18.47\%$ and $36.49\%$ in mAP, and $88.37\%$ and $96.21\%$ in mIoU on Oxford-5k and Paris-6k, respectively. Its performance gain over the Base is significant on Paris ($3.59\%$ in mAP). 
By contrast, it is marginal on Oxford-5k ($0.06\%$ in mAP).
The reason is that HAR is a technique designed for more accurate matching of instances, which means that it focuses more on the precision and the recall might be paid as the trade-off.
This can be seen from the slight drop of mAP@10 and mAP@20. It is an indication of the ranking decline of some top-ranked relevant targets in the Base.
With that being said, the mAP still shows improvement on both datasets. 
There must be a certain degree of precision rise which can compensate the recall decline and thus results in the improvement.
The compensation can be found from the gain in mIoU on Oxford-5k ($4.75\%$) in Table~\ref{tb:map}, and in more detail in Figure~\ref{fig:iou} where the superiority of Base+HAR over the Base is consistently observed.
Therefore, the degree of improvement depends on the degree of trade-off.
%

%
%This is more intuitive in the examples shown in Figure~X. The HAR has refined the attentions on both the queries and targets, and the regions are more regulated and better matched.

\begin{figure}
    \centering
    \includegraphics[width=0.95\columnwidth]{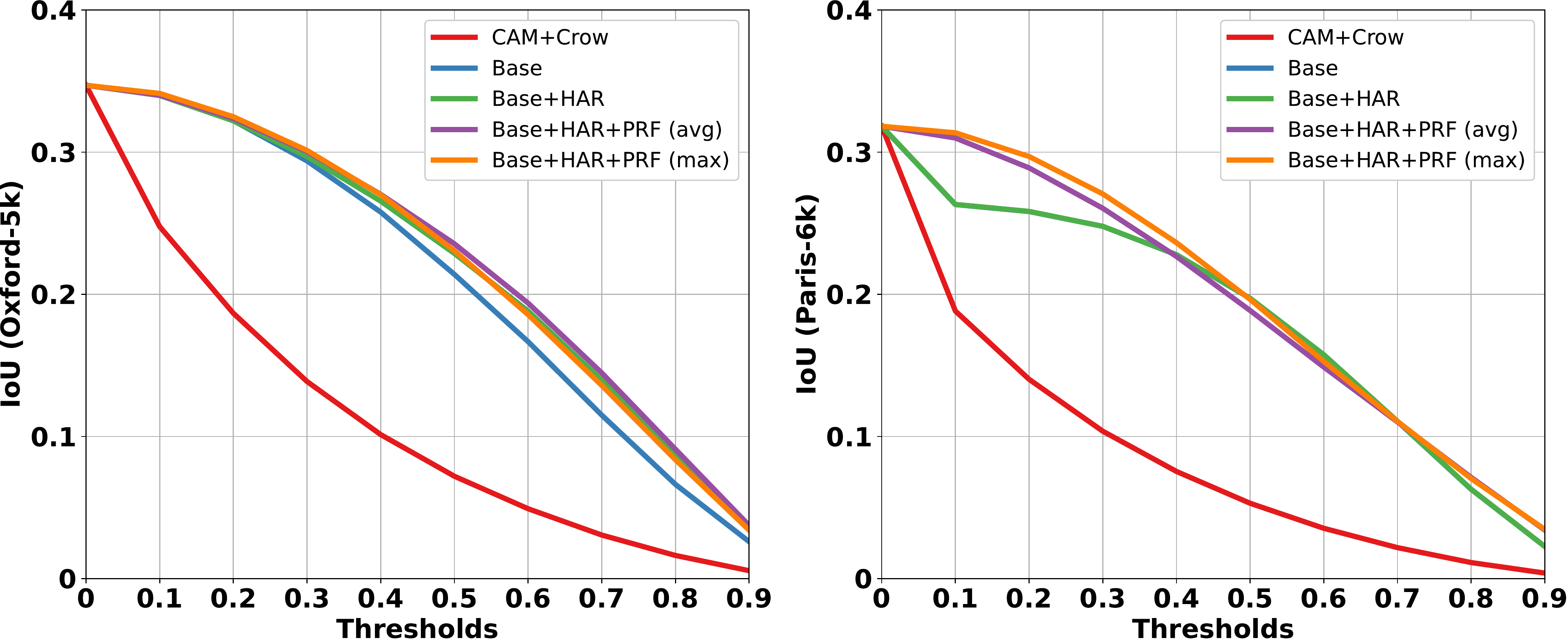}
    \caption{The IoUs over thresholds. The performance superiority of our methods over the region-specific CAM+Crow is observed.}
    \label{fig:iou}
\end{figure}

\subsection{Indicative Pseudo Relevance Feedback}
The performance by further integrating Indicative PRF is also evaluated. The improvement of Base+HAR+PRF ($\boldsymbol{\bar{\mathcal{Q}}}^*$) over Base+HAR has been observed on Paris-6k by $0.39\%$ and $4.88\%$ in mAP and mIoU, respectively. 
However, it causes a minor decline on Oxford-5k in mAP. This is again due to the trade-off between the precision and recall. The clue lies in Figure~\ref{fig:iou}, in which the Base+HAR+PRF runs clearly outperform other runs in IoU. 

Comparing between the \textit{average} and \textit{max} pooling, we can see in Figure~\ref{fig:iou} that the advantage of the \textit{max} before the threshold $0.4$ is more obvious than that after.
It is an indication that the improvement brought by the \textit{max} results from the IoU gain in recall, which means the contributions maps by the \textit{max} are more evenly distributed in the images than those by the \textit{average}.
In other words, those by the \textit{average} are more densely distributed on the highly confident regions, the (in-image) precision increases but not enough to compensate the decline of the recall.

\begin{table*}
	\centering \caption{Performance comparison to the state-of-the-art methods using region-specific attention. The \textbf{Ours} denotes our best runs of Base+$\boldsymbol{\Theta}^\gamma$ and Base+$\boldsymbol{\Theta}^\gamma$+$\boldsymbol{\bar{\mathcal{Q}}}^*$ on Oxford-5k and Paris-6k, respectively. The mAP@$k$ denotes the mAP evaluated through the top-$k$ returned results. The mIoU is the mean of IoUs over thresholds over [0.0 0.9]. The best results are in bold font.} \label{tb:sota}
	\setlength\extrarowheight{1.5pt}\setlength\tabcolsep{2pt}
	\renewcommand{\multirowsetup}{\centering}
	{\footnotesize%begin table fontsize
		\begin{tabular}{l|ccccc|ccccc}
			\hline
			& \multicolumn{5}{c|}{\small Oxford-5k} &\multicolumn{5}{c}{\small  Paris-6k}\\ 
			\cline{2-11}
			{\small Method}   & mAP & mAP@5  &mAP@10 &mAP@20  & mIoU  & mAP & mAP@5  &mAP@10 &mAP@20 & mIoU \\ \hline
			SPoC    & 84.44$\pm$3.24         &97.64$\pm$2.25        &94.36$\pm$3.36     &91.51$\pm$3.30         &14.71$\pm$9.01
			& 73.60$\pm$4.10      & 97.86$\pm$1.81           & 94.41$\pm$3.22           &91.71$\pm$3.63            &10.99$\pm$7.01\\
			Crow     &85.14$\pm$4.61         &98.12$\pm$2.86        &95.58$\pm$2.96      &91.97$\pm$3.48      &16.10$\pm$10.63
			&72.23$\pm$4.50   &97.28$\pm$2.76           & 94.31$\pm$3.59           &91.28$\pm$3.70            &12.19$\pm$8.18\\
			CAM+Crow &82.53$\pm$4.02  &95.43$\pm$3.79        &91.53$\pm$2.92      &88.49$\pm$2.82              &11.95$\pm$8.26 
			&71.38$\pm$4.83 & 96.54$\pm$3.75          & 93.61$\pm$3.70      &89.73$\pm$5.02            &9.51$\pm$9.33 \\
			Ours &\textbf{97.77$\pm$2.12}           &\textbf{100.00$\pm$0.00}      &\textbf{99.70$\pm$0.91}     &\textbf{99.49$\pm$1.47}            &\textbf{22.51$\pm$10.66} 
				&\textbf{97.81$\pm$0.90}           &\textbf{100.00$\pm$0.00}         &\textbf{99.95$\pm$0.10}          &\textbf{99.80$\pm$0.32}   &\textbf{19.57$\pm$10.17}\\
			\hline
		\end{tabular}
	}%set table fontsize
\end{table*}

\begin{figure}[t]
	\centering
	\includegraphics[width=0.48\textwidth]{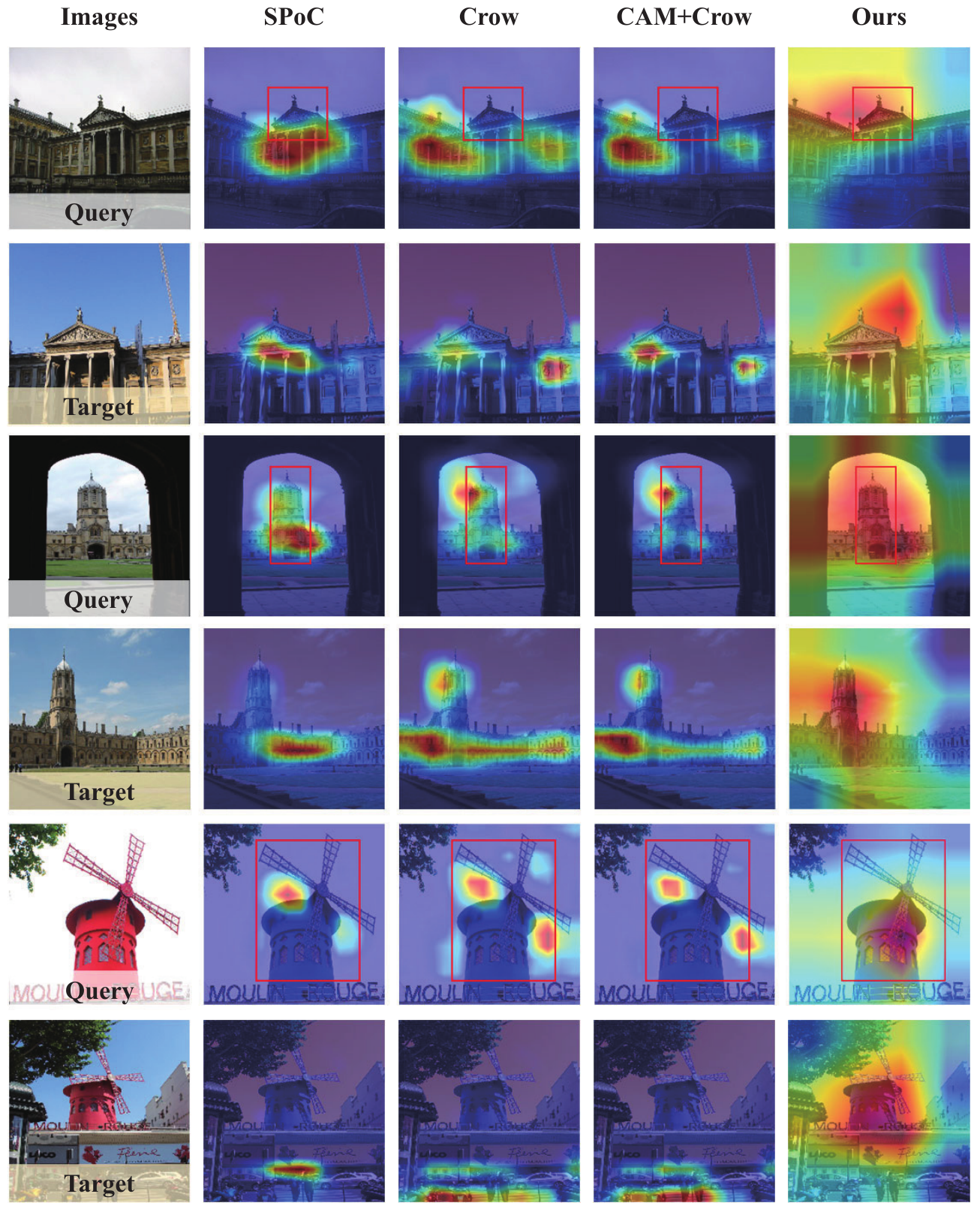}
	%\vspace{-0.1in}
	\caption{Comparison of the decoded contribution maps to the CAM results of SOTA. Our method shows more reasonable matched regions between the queries and targets.}\label{fig:sota_samples}
	%\vspace{-0.2in}
\end{figure}

\begin{figure}
    \centering
    \includegraphics[width=0.95\columnwidth]{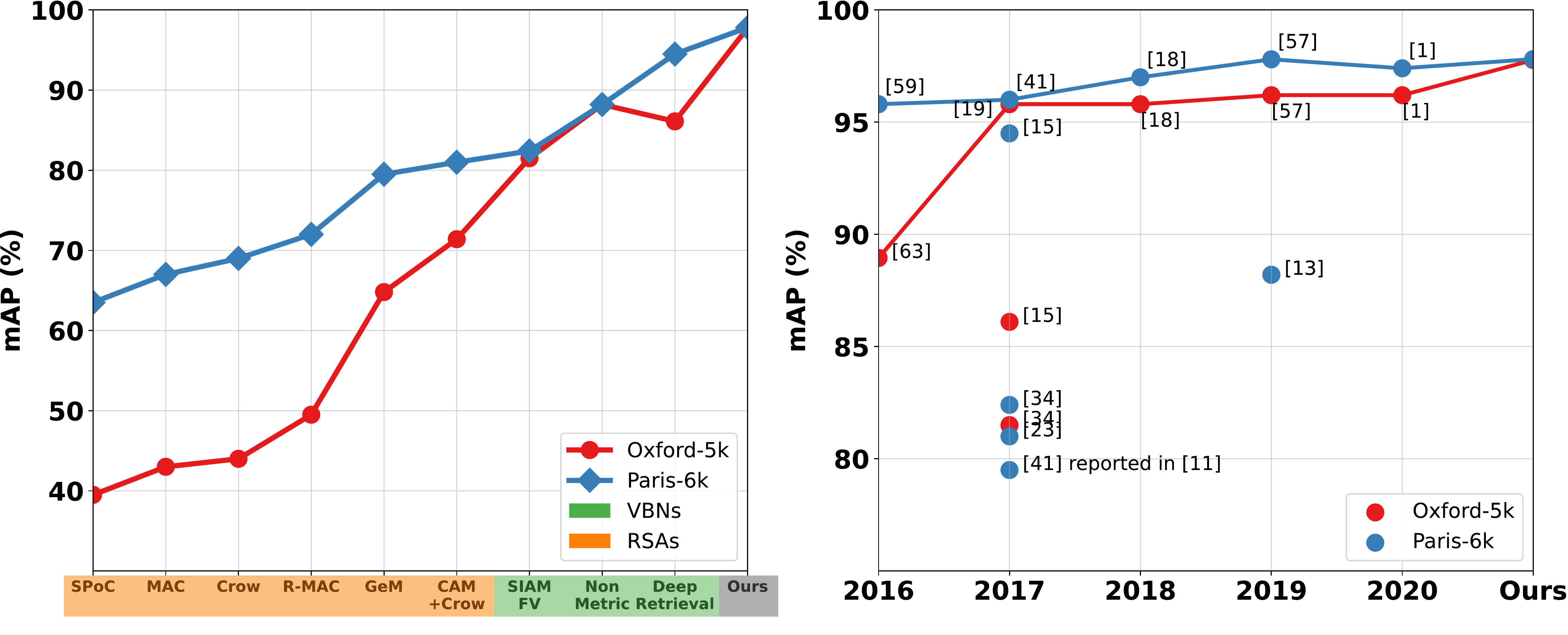}
    \caption{Performance comparison to the SOTA. Left: comparison to two groups methods (RSAs and VBNs) that are related to this paper. Right: comparison to methods reported with mAP greater than $0.75$ in recent 5 years (the best performance of each year is connected).
    Our method further set a new record of $97.77\%$ on Oxford-5k ($97.81\%$ on Paris).}
    \label{fig:map_comparison}
\end{figure}

\subsection{Comparison to the State-of-the-art (SOTA)}
\subsubsection{Fine-scale Comparison}
We conduct a fine-scale comparison to the regions-specific attention methods (RSAs) including SPoC \cite{babenko2015aggregating}, MAC \cite{razavian2016visual}, Crow \cite{kalantidis2016cross}, R-MAC \cite{tolias2015particular}, GeM \cite{radenovic2018fine}, and CAM+Crow \cite{jimenez2017class}, and the verification-based networks (VBNs) including SIAM-FV \cite{ong2017siamese}, Non-metric \cite{garcia2019learning}, and Deep Retrieval \cite{gordo2017end}.
However, the detailed mAPs and IoUs can only be reported for SPoC, CroW, and CAM+Crow, because of the availability of the source code and pre-trained models.
For the rest of methods, we adopt the results reported in the survey \cite{chen2021deep}.

The detailed comparison of mAPs and IoUs are shown in Table~\ref{tb:sota}. We can see our method gains the superiority over all region-specific methods in both mAP and mIoU. This is more intuitively shown in Figure~\ref{fig:sota_samples}. Our method generates more indicative evidence of the matching.

The comparison to the RSAs and VBNs is shown in Figure~\ref{fig:map_comparison} (Left). Our method performances the RSAs and VBNs by $66.99\%\pm 42.39$ and $12.91\%\pm 5.48$, respectively.
The VBNs generally obtain better performance over the RSAs. This again confirms the importance of matching modeling.

\subsubsection{Coarse-scale Comparison}
We also compare our method to 11 representative methods (only those with mAP greater than $75\%$ and developed in recent 5 years) including \cite{zheng2016accurate,iscen2017efficient,iscen2018fast,yang2019efficient,alemu2020multi,yang2016cross,radenovic2018fine,jimenez2017class,ong2017siamese,garcia2019learning,gordo2017end}. The results are shown in Figure~\ref{fig:map_comparison} (Right).
Our method obtains the best performance on both datasets among all methods in literature. It set a new record of mAP $97.77\%$ on Oxford-5k ($1.63\%$ better than the second best \cite{alemu2020multi,yang2019efficient}). This is not easy, given the fact that the performances being reported recently on this dataset are approaching optimal.
On Paris-6k, our method also wins the first place with mAP $97.81\%$ (the second best is $97.80\%$ \cite{yang2019efficient} reported in 2019). It also outperforms 10 methods in literate by $8.21\%\pm 8.81$.

%As our indicative retrieval focus on giving detailed evidence, this group is the most related. We have selected SPoC [], CroW [], and CAM+Crow [], as their performance are widely recognized in literature.
    
%Verification-based Retrieval: As our method can be also considered as a verification-based scheme, we have selected the AP Loss [] as the representative, because it optimizes the AP directly and is with promising performance.

\section{Conclusion}
We conduct a pilot study for revisiting the importance of matching modeling for indicative image retrieval.
By skipping the representation learning, our framework starts from the explicit correspondence modeling directly, and thus removes the dependency on pre-trained models and related issues.
%
%With the explicit correspondence modeling, matched pairs can be backtracked for the indicative evidence generation.
%
It also gives us the space for conducting explicit attention regulation and pseudo-relevance feedback.
More importantly, we have reclaimed the significance of matching modeling by demonstrating that promising performance can be obtained without extracting any deep features. 

While encouraging results being obtained, there are still some limitations. To name a few, the variance of image changes is not considered (e.g., the scale, illumination, and rotation changes), and the indexing is not easy to conduct because of the verification-based setting.
However, the indicative image retrieval is an open framework to integrate other retrieval techniques.
Sophisticated techniques can be integrated to address problems not covered in this study.
It is also not necessarily being exclusive to feature extraction and representation learning. 

{\small
\bibliographystyle{ieee_fullname}
\bibliography{cvpr}
}
\end{twocolumn}
\end{document}